\documentclass[runningheads]{llncs}
\usepackage{graphicx}
\usepackage{tikz}
\usepackage{comment}
\usepackage{amsmath,amssymb} % define this before the line numbering.

\usepackage{color}
\usepackage{graphicx}
\begin{document}
% \linenumbers
\pagestyle{headings}
\mainmatter
\def\ECCVSubNumber{100}  % Insert your submission number here

\title{Probabilistic Deep Learning for Instance Segmentation} % Replace with your title

%\begin{comment}
%\titlerunning{ECCV-20 submission ID \ECCVSubNumber}
%\authorrunning{ECCV-20 submission ID \ECCVSubNumber} 
\author{Anonymous ECCV submission}
\institute{}% Paper ID \ECCVSubNumber
%\end{comment}
%******************

%\begin{comment}
\titlerunning{Probabilistic Instance Segmentation}
\author{Josef Lorenz Rumberger\inst{2,3} \and
Lisa Mais\inst{1,3} \and
Dagmar Kainmueller\inst{1,3}}
\authorrunning{Rumberger et al.}
% I wrote down all institutes, feel free to delete some :)
\institute{
Berlin Institute of Health, Berlin, Germany
\and
Charité University Hospital, Berlin, Germany
\and
Max Delbr\"uck Center for Molecular Medicine, Berlin, Germany \\
\email{\{joseflorenz.rumberger,lisa.mais,dagmar.kainmueller\}@mdc-berlin.de}}
%\end{comment}
%******************
\maketitle

\begin{abstract}
Probabilistic convolutional neural networks, which predict distributions of predictions instead of point estimates, led to recent advances in many areas of computer vision, from image reconstruction to semantic segmentation. Besides state of the art benchmark results, these networks made it possible to quantify local uncertainties in the predictions. These were used in active learning frameworks to target the labeling efforts of specialist annotators or to assess the quality of a prediction in a safety-critical environment. However, for instance segmentation problems these methods are not frequently used so far. We seek to close this gap by proposing a generic method to obtain model-inherent uncertainty estimates within proposal-free instance segmentation models. Furthermore, we analyze the quality of the uncertainty estimates with a metric adapted from semantic segmentation. We evaluate our method on the BBBC010 C.\ elegans dataset, where it yields competitive performance while also predicting uncertainty estimates that carry information about object-level inaccuracies like false splits and false merges. We perform a simulation to show the potential use of such uncertainty estimates in guided proofreading. 
\end{abstract}

\section{Introduction}
Probabilistic deep learning models predict distributions of predictions instead of single point estimates and make it possible to quantify the inherent uncertainty of predictions. They were successfully applied for computer vision tasks such as image classification \cite{blundell2015weight,kingma2015variational,gal2016dropout,gal2017concrete,gal2017deep}, semantic segmentation \cite{mukhoti2018evaluating,kendall2017uncertainties,gal2017concrete} and regression problems like instance counting \cite{oh2020crowd} and depth regression\cite{kendall2017uncertainties}. They achieve state-of-the-art results for complex problems such as predicting distributions for segmentation tasks that carry inherent ambiguities \cite{kohl2018probabilistic} or microscopy image restoration \cite{weigert2018content}. 
However, said probabilistic deep learning methods are not directly applicable to proposal-free instance segmentation approaches, which define the current state-of-the-art in some applications from the bio-medical domain~\cite{kulikov2020harmonic,hirsch2020patchperpix,funke2018large}. This is due to the fact that proposal-free instance segmentation methods require some form of inference to yield instance segmentations from network predictions. Thus, to the best of our knowledge, probabilistic deep learning models have not yet been studied in the context of proposal-free instance segmentation methods.
To close the gap, this work (1) makes use of a probabilistic CNN to estimate the local uncertainty of a metric-learning based instance segmentation model, and (2) examines these estimates with regard to their informativeness on local inaccuracies. 
On the challenging BBBC010 C. elegans dataset~\cite{wahlby2012image}, our model achieves competitive performance in terms of accuracy, while additionally providing estimates of object-level inaccuracies like false splits or false merges.

The concept of sampling hypotheses from a probabilistic model for instance segmentation was already used in~\cite{funke2014candidate}. However, their work differs in that the candidates are not obtained from an end-to-end trainable model as well as they do not consider uncertainties.
The idea of predicting local uncertainty estimates for instance segmentation tasks has been proposed by \cite{Morrision2019ProbMaskCNN}, who use dropout sampling on a Mask-RCNN \cite{he2017mask}. However, their model employs a proposal-based segmentation method, which carries the critical disadvantage that bounding-boxes are required to be sufficient region proposals for objects. For thin, long and curvy structures  which often arise in the bio-medical domain, bounding-boxes frequently contain large parts of instances of the same category, which deteriorates segmentation performance \cite{fathi2017semantic}. Furthermore, \cite{Morrision2019ProbMaskCNN} limit their analysis of results to a probabilistic object detection benchmark \cite{hall2020probabilistic}, while we quantitatively assess the quality of the uncertainty estimates for instance segmentation. 

Overcoming the limitations of proposal-based methods, state-of-the-art \newline proposal-free models use a CNN to learn a representation of the data that allows instances to be separated. Popular approaches are based on learning and post-processing a watershed energy map \cite{bai2017deep,wolf2017learned}, an affinity-graph \cite{funke2018large,hirsch2020patchperpix,liu2018affinity,wolf2018mutex} or a metric space \cite{de2017semantic,fathi2017semantic} into binary maps for each instance. However, this binary output misses information like confidence scores or uncertainty measures. 

In Bayesian machine learning, at least two kinds of uncertainties are distinguished that together make up the predictive uncertainty: Data uncertainty (i.e. aleatoric uncertainty) accounts for uncertainty in the predictions due to noise in the observation and measurement process of data or ambiguities in the annotation process and thus does not decrease with more training data \cite{kendall2017uncertainties}. 
Model uncertainty (i.e. epistemic uncertainty) captures the uncertainty about the model architecture and parameters. To account for it, parameters are modeled as probability distributions. The more data gets acquired and used for training, the more precise one can estimate their distribution and the smaller the variance becomes. Thus, model uncertainty decreases with more data. This work focuses on the estimation of model uncertainty, since it is high in applications with small annotated datasets and sparse samples \cite{kendall2017uncertainties}, which is typical for bio-medical image data.

Model uncertainty is commonly estimated by approximating the unknown parameter distributions by simple variational distributions. A popular choice is the Gaussian distribution, which has proven effective but leads to high computational complexity and memory consumption~\cite{blundell2015weight,kingma2013auto}. To overcome these limitations, \cite{kingma2015variational,gal2016dropout} proposed to use Bernoulli distributions instead, which can be implemented via Dropout. 
For our work, we use the Concrete Dropout model \cite{gal2017concrete}, because it 
has shown high quality uncertainty estimates for semantic segmentation tasks~\cite{mukhoti2018evaluating} and provides learnable dropout rates due to its variational interpretation. To our knowledge it has not been applied to instance segmentation tasks yet.

In order to obtain draws from the posterior predictive distribution in the Concrete Dropout model~\cite{gal2017concrete}, a single input image is passed several times through the network, each time with a new realization of model parameters. Our proposed pipeline post-processes each such draw into binary instance maps, and agglomerates the resulting predictions into a single probabilistic prediction for each instance.
Our proposed pipeline is constructed to be model-agnostic and applicable to any CNN-based proposal- free instance segmentation method. For showcasing, in this work, we pick a metric learning model, because (1) post-processing is fast and simple, and (2) metric learning models have shown competitive performance on several challenging datasets~\cite{de2017semantic,luther2019learning}.

In summary, the key contributions of this work are:
\begin{itemize}
    \item To the best of our knowledge, this is the first work that uses a Bayesian approximate CNN in conjunction with proposal-free instance segmentation.
    \item We adapt a metric for quantitative comparison of different uncertainty estimates  \cite{mukhoti2018evaluating}, originally proposed for semantic segmentation, to the case of instance segmentation.
\end{itemize}
\section{Methodology} \label{sec:method}
This section describes the models employed in our proposed pipeline, as well as the associated loss functions and post-processing steps. Furthermore, we propose an adaptation of an uncertainty evaluation metric for probabilistic semantic segmentation~\cite{mukhoti2018evaluating} to the case of instance segmentation.

\subsection{Models and Losses} 
\subsubsection{Metric Learning with the Discriminative Loss Function:}
We follow the proposal-free instance segmentation approach of~\cite{de2017semantic}, a metric learning method which predicts, for each pixel, a vector in an embedding space, and trains for embedding vectors that belong to the same instance to be close to their mean, while mean embeddings of different instances are trained to be far apart. This is achieved by means of the \emph{discriminative loss function}, which consists of three terms that are jointly optimized: The variance term \eqref{eq:disc_var} pulls embeddings $e_{c,i}$ towards their instance center $\mu_c$ in the embedding space: 
\begin{equation} \label{eq:disc_var}
    L_{var} = \frac{1}{C}\sum\limits^C_{c=1} \frac{1}{N_c}\sum\limits^{N_c}_{i=1} ||\mu_c-e_{c,i} ||^2
\end{equation}
with $C$ the total number of instances and $N_c$ the total number of pixels of instance $c$. The original formulation in \cite{de2017semantic} included a hinge that set the loss to zero for embeddings that are sufficiently near the center. However, in our work we use the version proposed by \cite{luther2019learning}, which excludes the hinge in order to have lower intra-cluster variance of embeddings which is desirable to prevent false splits during post-processing. The distance term \eqref{eq:disc_dist} penalizes cluster centers $c_A$  
\begin{equation} \label{eq:disc_dist}
    L_{dist} = \frac{1}{C(C-1)}\sum\limits^C_{c_A=1}\sum\limits^C_{\substack{c_B=1\\c_A\neq c_B}}[2\delta_d - ||\mu_{c_A}-\mu_{c_B}||^2]^2_+
\end{equation}
and $c_B$ for lying closer together than $2\delta_d$ and therefore pushes clusters away from each other. Distance hinge parameter $\delta_d = 4$ is used to push clusters sufficiently far apart. Choosing this hyperparameter posed a trade-off: a higher $\delta_d$ led to a wider separation of clusters and increased segmentation performance but also made the loss have high jumps between samples, which led to training instability.

The last part is the regularization term \eqref{eq:disc_reg}, which penalizes the absolute sum of the embedding centers and therefore draws them towards the origin. 
 \begin{equation}\label{eq:disc_reg}
    L_{reg} = \frac{1}{C}\sum\limits^C_{c=1}||\mu_c||^2\\
\end{equation}
In order to jointly optimize, the terms are weighted as follows: $ L_{disc} =  L_{var} + L_{dist} + 0.001 \cdot L_{reg}$. Besides this loss, a three-class cross-entropy loss function is added in order to learn to distinguish background, foreground and pixels that belong to overlapping instances.

\subsubsection{Baseline Model:}
We employ a U-Net~\cite{ronneberger2015u} as backbone architecture, which is a popular choice for pixel-wise prediction tasks~\cite{funke2018large,hirsch2020patchperpix,luther2019learning}. The network is trained with weight decay to make it comparable to our Concrete Dropout Model described in the following.

\subsubsection{Concrete Dropout Model:}
We employ the Concrete Dropout model proposed in~\cite{gal2017concrete}. The remainder of this Section is our attempt to motivate and describe this model to the unfamiliar reader by means of intuitions (where possible) and technical details (where necessary). It does not intend to serve as comprehensive summary of the respective theory, for which we refer the reader to~\cite{gal2017concrete}.

It has been shown that Dropout, when combined with weight decay, can be interpreted as a method for approximating the posterior of a deep Gaussian Process~\cite{gal2016dropout}. This finding provides a theoretical basis to the practical approach of assessing output uncertainties from multiple predictions with Dropout at test time. 

An intuitive detail of the respective theory is that a single weight matrix $M \in \mathbf{R}^{k\times l}$, together with a dropout rate $p$, implements a distribution over weight matrices of the form $M\cdot \textnormal{diag}(\left[z\right]_{i=1}^{k})$, with $z_i\sim\textnormal{Bernoulli}(1-p)$.
This, however, entails that for fixed dropout rates, output uncertainty scales with weight magnitude. This is not desired, as high magnitude weights may be necessary to explain the data well. This discrepancy, formally due to a lack of $calibration$ of Gaussian Processes, motivates the need for dropout rates to be learnt from data, simultaneously with respective weight matrices~\cite{gal2017concrete}. 

A respective training objective has been proposed in~\cite{gal2017concrete}(cf.\ Eq.\ 1 -- 4 therein). The objective stems, as in~\cite{gal2016dropout}, from the idea of approximating the posterior of a deep Gaussian Process, with the difference that~\cite{gal2017concrete} consider learnable dropout rates. The objective combines a data term, which pulls dropout rates towards zero, and boils down to a standard SSD loss for regression problems, with a regularizer, which keeps the joint distribution over all weight matrices close to the prior of a Gaussian Process. This regularizer effectively pulls dropout rates towards 0.5 (i.e. maximum entropy), and is weighted by one over the number of training samples available, which entails higher dropout rates / higher uncertainty for smaller training set size. More specifically, the regularizer takes the following form (see Eq.\ 3,4 in~\cite{gal2017concrete}):
\begin{align} 
    L_{concrete} &= \frac{1}{N}\cdot\left(\sum\limits_{l=1}^L \frac{\iota^2(1-p_l)}{2} ||M_l||^2 - \zeta F_l \mathcal{H}(p_l)\right) \label{eq:concrete_loss} \\
    \textnormal{with } \mathcal{H}(p_l)&:= -p_l ln(p_l) - (1-p_l)ln(1-p_l) \label{eq:entropy}
\end{align}
The first part of the function is an $l_2$ regularizer on the pre-dropout weight matrices $M_l$, with $l$ denoting one of the $L$ layers of the network, $p_l$ the dropout probability of the respective layer, $N$ as the number of training examples, and $\iota$ the prior length scale, which is treated as a hyperparameter that controls the strength of the regularizer. 
The second term serves as a dropout rate regularizer. It captures the entropy $\mathcal{H}(p_l)$ of a Bernoulli random variable with probability $p_l$, where high entropy is rewarded. Hyperparameter $\zeta$ serves as weight for this term. It is furthermore scaled by the number of nodes per layer $F_l$, thus encouraging higher dropout rates for layers with more nodes. Note that hyperparameter $\iota$ gauges a weight decay effect implemented by this loss. Thus we set it equal to the weight decay factor in our baseline model to yield comparable effects.

To form the loss for our Concrete Dropout Metric Learning model, we add $L_{concrete}$ to the discriminative loss from our baseline metric learning objective, which is a weighted sum of terms \ref{eq:disc_var}, \ref{eq:disc_dist}, and \ref{eq:disc_reg}. Note that term~\ref{eq:disc_var} is a sum of squared differences loss and hence constitutes a theoretically sound data term in the Concrete Dropout objective. It is, however, unclear if terms \ref{eq:disc_dist} and \ref{eq:disc_reg} can be grounded in the same theoretical framework, and a respective analysis is subject to future work.

Our loss has the drawback that it is difficult to optimize w.r.t.\ dropout rates. (Note that each of its four terms depends on the dropout rates.) The Concrete Dropout Model~\cite{gal2017concrete} alleviates this drawback by interpreting each (layer-individual) dropout rate as parameter of a Concrete distribution, which is a continuous relaxation of the respective Bernoulli distribution. 

\subsection{Post-Processing}  
To yield an instance segmentation from predictions of embeddings, mean-shift clustering  \cite{fukunaga1975estimation} is applied to find cluster centers in the embedding space.
All embeddings within a given threshold of a cluster center are gathered and their corresponding pixels represent an instance. 
In~\cite{de2017semantic}, the authors propose to use the hinge parameter from the variance term $L_{var}$ as the mean-shift bandwidth and the clustering threshold during post-processing. In the version of the loss used here, eq.~\eqref{eq:disc_var} does not have a hinge parameter. Therefore, the mean-shift bandwidth and the clustering threshold are treated as hyperparameters and found through a grid-search with 2-fold cross-validation on the test set. To reduce the number of embeddings considered during clustering, embeddings at pixels classified as background are ignored. 
To tackle overlapping instances, we propose a  straightforward heuristic: For each pixel, background/foreground/overlap probabilites are predicted in addition to embeddings. We add the resulting overlap map, containing all overlaps of a sample, to each individual instance map. In each instance-plus-overlap map, the connected component with the largest Intersection-over-Union (IoU) with the respective sole instance map is selected as the final instance prediction.

\subsection{Inference on the Probabilistic Model} 
Every sample is passed eight times through the Concrete Dropout U-Net to obtain draws from the posterior predictive distribution. Each draw is post-processed individually by mean-shift clustering, such that eight instance segmentations are generated.

We propose a straightforward approach to capture uncertainties on the instance level: Each instance contained in a single draw is first converted into an individual binary segmentation. The draw that contains the highest total number of instances is the base of the following agglomeration. By choosing this draw as the first base map, the agglomeration strategy decreases false negatives in the final prediction. A second draw of instance segmentation maps is taken 
and a linear assignment problem is constructed by calculating the IoU for every pair of instance segmentations that belong to the different draws. To solve this, the Kuhn-Mukres (i.e. Hungarian) matching algorithm \cite{munkres1957algorithms} is employed, which finds a mapping from the base set of instances to the other, such that the sum of IoUs is maximized.
To aggregate more draws, a union of the two maps is calculated and taken as the base for subsequent steps. When all draws are agglomerated, the summed up instance maps are divided by the number of draws from the posterior to obtain a probability for every pixel to belong to a given instance. 

\subsection{Uncertainty Evaluation}
In order to quantitatively assess the quality of our uncertainty estimates, we adapt the metrics presented by Mukhoti and Gal \cite{mukhoti2018evaluating} to the task of instance segmentation. The authors state that good uncertainty estimates should be high where the model is inaccurate and low where it is accurate, and formalize this into the following conditional probabilities:
\begin{enumerate}
    \item $p(accurate|certain)$: the probability that a models predictions are accurate given it is certain about it.
    \item $p(uncertain|inaccurate)$: the probability that a model is uncertain, given its prediction is inaccurate.
\end{enumerate}
The authors use the metrics on $2 \times 2$ image patches and calculate the mean accuracy and uncertainty of each patch. Then both mean values are converted into binary variables by thresholding the accuracy with $0.5$ and plotting the probabilities as a function of uncertainty thresholds. Here, these metrics are adapted for instance segmentation by focusing on the precision instead of the accuracy. We do so, by excluding all image patches that neither belong to the foreground in the prediction nor the groundtruth and calculate the following for every instance prediction:
\begin{align} \label{eq:cond_probs}
    &p(accurate|certain) = \frac{n_{ac}}{(n_{ac}+n_{ic})} \\
    &p(uncertain|inaccurate) = \frac{n_{iu}}{(n_{ic}+n_{iu})}
\end{align}
With $n_{ac}$ the number of patches that are accurate and certain, $n_{iu}$ the patches that were inaccurate and uncertain and the two undesired cases (accurate and uncertain, inaccurate and certain). Both probabilities are then combined into a single metric, called Patch Accuracy vs. Patch Uncertainty (PAvPU) \cite{mukhoti2018evaluating}:
\begin{equation}
    PAvPU = \frac{(n_{ac} + n_{iu})}{(n_{ac} + n_{au} + n_{ic} + n_{iu})}
\end{equation}
Since instance segmentation pipelines return at least binary maps for every instance, we interpret a single segmentation of an instance as a draw from a Bernoulli random variable and use the entropy of the Bernoulli as stated in eq.~\eqref{eq:entropy} as the uncertainty measure~\cite{mukhoti2018evaluating}. Instead of $2 \times 2$ image patches, we use $4 \times 4$ patches for the calculation of the metrics, since we are more interested in larger structural errors and less in small boundary adherence errors.
\section{Results}
This section presents benchmark results of the proposed methods on the wormbodies BBBC010 dataset \cite{wahlby2012image} and an analysis of the uncertainty estimates. 
Both models employ a 5-level U-Net architecture with same-padding, ReLU activation functions in all hidden layers and filter size $(3,3)$. The number of filters increases from $19$ in the first layer with each down-sampling step by factor $2$ and decreases vice versa for up-sampling steps. The $l_2$ weight decay is weighted with factor $1\mathrm{e}{-6}$ for the baseline model and the corresponding parameter in the concrete dropout model is set to $\iota^2 = 1\mathrm{e}{-6}$ to get a comparable regularization. The dropout rate regularizer hyperparameter is set to $\zeta=1\mathrm{e}{-3}$ which leads to dropout rates up to $50\%$ in deep layers of the U-Net and near zero rates for in- and output layers. 
Models are trained for $800,000$ iterations on random $512\times 512$ pixel sized slices of the data to predict 16 dimensional embedding vectors. Standard data augmentations, including random rotations and elastic deformations, are applied. The dataset is split into a train and a test set with 50 samples each as in \cite{hirsch2020patchperpix,novotny2018semi,yurchenko2017parsing} and we perform 2-fold cross-validation to determine hyperparameters. 

\setlength{\tabcolsep}{3pt}
\begin{table}[]
\begin{center}
\caption{Evaluation on the BBBC010 test set and comparison to state of the art. Especially our Concrete Dropout model yields competitive performance while also predicting uncertainty estimates that carry information about object-level inaccuracies. Best results are shown in bold, second best are underlined.}
\label{tab:benchmarks}
\begin{tabular}{lccccc}
\noalign{\smallskip}
\hline
\noalign{\smallskip}
 Models & avAP$_{[0.5:0.95]}$ & AP$_{.5}$ & AP$_{.75}$ & Recall$_{.8}$ & avAP$_{dsb[0.5:0.95]}$\\
\noalign{\smallskip}
\hline
\noalign{\smallskip}
Semi-conv Ops \cite{novotny2018semi}& 0.569 & 0.885 & 0.661  & -         & -\\
SON \cite{yurchenko2017parsing}& -     & -     & -      & $\sim$0.7  & - \\
Discrim. loss (from \cite{kulikov2020harmonic})& 0.343 & 0.624 & 0.380 & -  & - \\
Harmonic embed. \cite{kulikov2020harmonic} & 0.724 & 0.900 & 0.723  & -      & - \\
PatchPerPix \cite{hirsch2020patchperpix} & \textbf{0.775} & 0.939 & \textbf{0.891}  & \textbf{0.895} & \textbf{0.727}\\
\noalign{\smallskip}
\hline
\noalign{\smallskip}
Baseline model& 0.761 & \underline{0.963} & 0.879  & \underline{0.81}      &  0.686 \\
Concrete dropout model& \underline{0.770} & \textbf{0.974} & \underline{0.883}  & \underline{0.81}  & \underline{0.703} \\
\hline
\end{tabular}
\end{center}
\end{table}
\setlength{\tabcolsep}{1.4pt}

Table \ref{tab:benchmarks} shows benchmark results against state of the art models. Since the benchmark metrics require binary instance segmentation maps, the probabilities that the Concrete U-Net pipeline predicts are binarized with threshold 0.75. The presented average precision (AP, avAP) scores are the widely used MS COCO evaluation metrics \cite{lin2014microsoft}. As a complement, avAP$_{dsb[0.5:0.95]}$ follows the Kaggle 2018 data science bowl definition for AP scores ($AP_{dsb}=\frac{TP}{TP+FP+FN}$) that also accounts for false negatives. 

The concrete dropout model performs in all metrics slightly better than the baseline model, as expected based on \cite{gal2017concrete,mukhoti2018evaluating}. 
Both models outperform the competitors semi-convolutional operators \cite{novotny2018semi}, singling-out networks (SON) \cite{yurchenko2017parsing} and Harmonic Embeddings~\cite{kulikov2020harmonic} by a large margin. Especially interesting is the comparison with the recent Harmonic Embedding model~\cite{kulikov2020harmonic}. In their evaluation on BBBC010, \cite{kulikov2020harmonic} furthermore compared their model to a vanilla discriminative loss model, which yielded considerably lower performance than our models. The main differences of our baseline model when compared to the vanilla discriminative loss model employed in~\cite{kulikov2020harmonic} are that we perform hyperparameter optimization on the mean-shift bandwidth and clustering threshold, and don't employ a hinge parameter in Eq.\ \ref{eq:disc_var}. 
That said, both of our models show slightly lower performance than the recent PatchPerPix~\cite{hirsch2020patchperpix} except for the AP50 score, in which it constitutes the new state of the art. 

\begin{figure}[htb!]
\includegraphics[width=.95\textwidth]{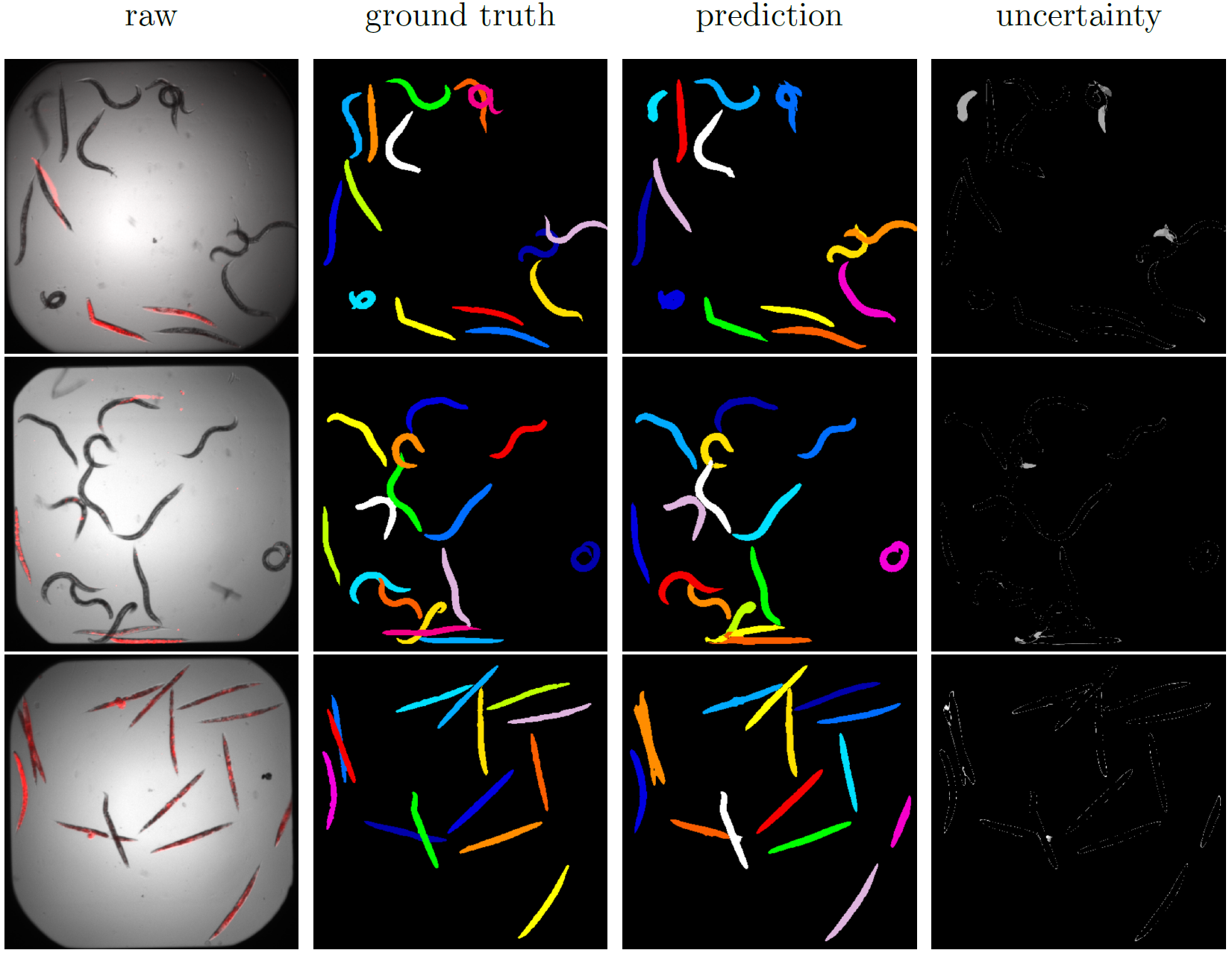}
\caption{Test set samples 'C10', 'C04' and 'D21', from top to bottom. SYTOX staining in red in the raw input images, predictions and associated uncertainties from the Concrete Dropout model}\label{fig:sample_predictions} % 
\end{figure}

\begin{figure} [htb!]
\centering
  \includegraphics[width=.95\textwidth]{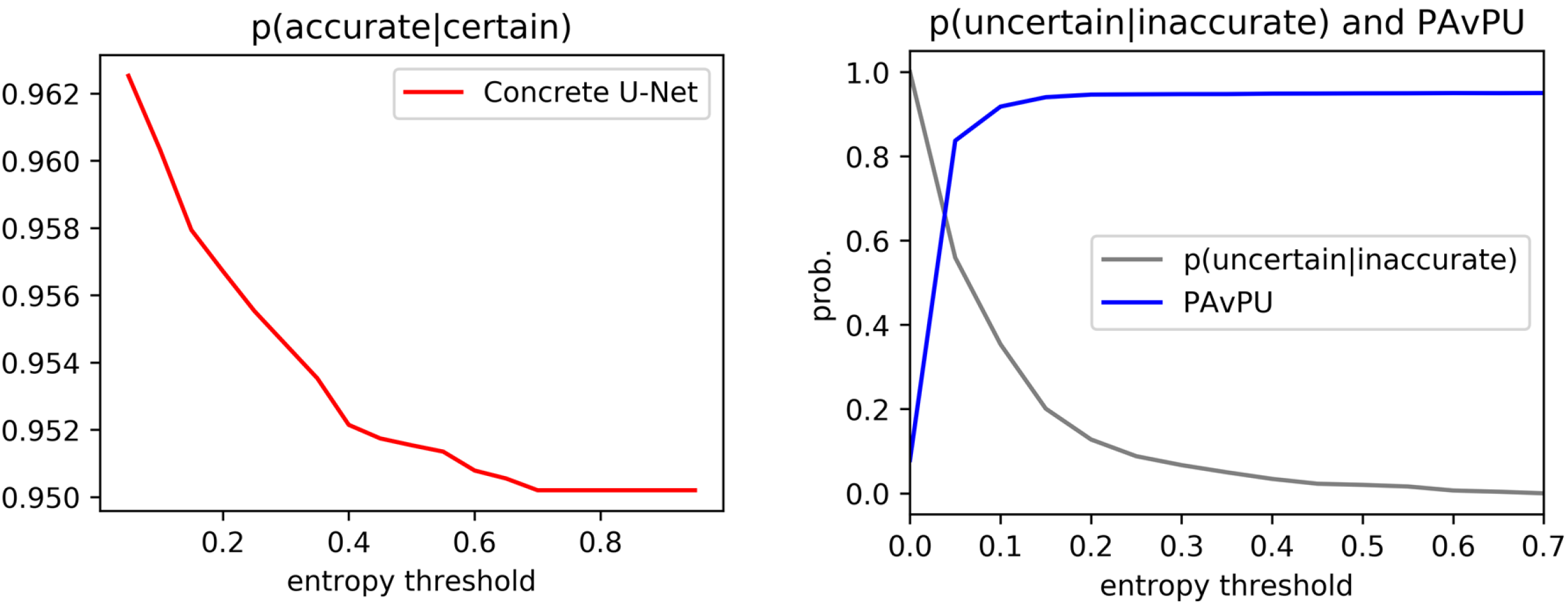}
\caption{Uncertainty metrics for test set predictions of the Concrete Dropout model} %
\label{fig:unc_metrics}
\end{figure}

Figure \ref{fig:sample_predictions} shows samples from the test set with associated predictions from the Concrete U-Net. Typical error cases of the predictions are false merges and incomplete segmentations, that occur where two or more worms overlap each other. The latter error also happens where worms are blurry due to movement or poor focus. At first glance, the uncertainty estimates shown in Figure \ref{fig:sample_predictions} are informative on the locations of segmentation errors.
A quantitative analysis of (un)certainty vs.\ (in)accuracy is presented in Figure \ref{fig:unc_metrics}. It shows the uncertainty metrics plotted against a raising entropy threshold. The smaller the entropy threshold, the more patches have an entropy exceeding the threshold and are thus denoted as uncertain. For threshold $0.05$, the probability that a patch whose entropy is below 0.05 is accurate is 0.962 (red curve). The probability that a given inaccurate patch is classified as uncertain at the 0.05 entropy threshold is 0.559 (gray curve). Therefore not all inaccurate patches can be targeted with this approach, even on this low threshold. 

\subsection{Simulation Experiment} \label{sec:simulation}
% Simulation experiment
One potential use-case for the uncertainty estimates is proof-reading guidance. In order to assess its feasibility, the following simulation experiment is conducted: For the $\{5,10,15,20\}$ highest uncertainty peak patches among all samples of the dataset, instance predictions that lie within or directly neighbor the peak patches are corrected.
In this simulation, correction is done by swapping the respective instance maps with their ground truth counterparts. Benchmark metrics for this simulation are shown in Table \ref{tab:simulation}.
\setlength{\tabcolsep}{3pt}
\begin{table}[t!]
\begin{center}
\caption{Simulation results. Number of corrections denotes the number of uncertainty peak patches that were used for error correction guidance. Concrete U-Net probabilities are binarized by thresholding with 0.75 to calculate the metrics}\label{tab:simulation}
\begin{tabular}{lccccc}
\noalign{\smallskip}
\hline
\noalign{\smallskip}
 Corrections & avAP$_{[0.5:0.95]}$ & AP$_{.5}$ & AP$_{.75}$ & Recall$_{.8}$  & avAP$_{dsb[0.5:0.95]}$\\
\noalign{\smallskip}
\hline
\noalign{\smallskip}
0 & 0.770 & 0.974 & 0.883  & 0.81     & 0.703 \\
5 & 0.779 & 0.977 & 0.883  & 0.82     & 0.714 \\
10 & 0.782 & 0.978 & 0.896  & 0.83    & 0.719 \\
15 & 0.786 & 0.980 & 0.897  & 0.83    & 0.725 \\
20 & 0.791 & 0.982 & 0.903  & 0.84    & 0.735 \\
\hline
\end{tabular}
\end{center}
\end{table}
\setlength{\tabcolsep}{1.4pt}
With increasing number of simulated corrections, the evaluation metrics raise substantially when considering that corrections are only done on one or two instances at a time. This shows that the uncertainty estimates carry enough information to guide manual proof-readers towards segmentation errors, and that a significant boost in segmentation quality can be obtained by means of a small number of targeted inspections.
\section{Discussion}
The proposed models show high benchmark results when compared to other methods. The difference between our models and the discriminative loss model in~\cite{kulikov2020harmonic} suggests that further performance gains could be reached by incorporating estimates for the mean-shift bandwidth and the clustering threshold into the model optimization objective. During hyperparameter search, we also observed that a smaller mean-shift bandwidth could be vital to prevent false-merges in some instance pairs, whereas it led to false-splits in others. This leads to the hypothesis, that an instance specific bandwidth could further boost performance.

Regarding the concrete dropout model, it shows slightly higher performance and its uncertainty estimates are informative on local inaccuracies. Nevertheless, there still exist many regions in the predictions that have both, segmentation errors and low local uncertainty estimates. To get more informative uncertainty estimates, a loss function that incorporates the quantification of data uncertainty \cite{oh2018modeling} into the model could be used instead of the discriminative loss function. Furthermore, Variational Inference techniques like Concrete Dropout have the drawback that their approximate posterior is just a local approximation of the full posterior distribution. Therefore global features of the posterior are neglected~\cite{murphy2012machine}, which reduces the quality of the uncertainty estimates. One could improve on that by using an ensemble of probabilistic models~\cite{smith2018understanding}. The models of the ensemble explore various local optima in the loss landscape and therefore various regions of the posterior distribution. Thus, the ensemble better reflects global features of the posterior, while still approximating the local features reasonably well. The simulation experiment intended to show a possible use case of the proposed method for practitioners. Other use cases are imaginable, like Bayesian active learning~\cite{gal2017deep}, which ranks unlabeled samples based on their uncertainty estimates to point the annotation efforts towards more informative samples during data generation.
\section{Conclusion}
We presented a practical method for uncertainty quantification in the context of proposal-free instance segmentation. We adapted a metric that evaluates uncertainty quality from semantic to instance segmentation. Furthermore, we also adapted a metric learning method to be able to cope with overlapping instances. This work is just a first step towards a probabilistic interpretation of instance segmentation methods and an important future research topic is the formulation of a loss function that incorporates data related uncertainty estimates. 

\vspace{2em}
\noindent\textbf{Acknowledgments. }
J.R. was funded by the German Research Foundation DFG RTG 2424 \emph{CompCancer}. L.M. and D.K. were funded by the Berlin Institute of Health and the Max Delbrueck Center for Molecular Medicine and were supported by the HHMI Janelia Visiting Scientist Program.

\clearpage
\bibliographystyle{splncs04}
\bibliography{egbib}
\end{document}